\documentclass{article}

\usepackage[final,nonatbib]{nips_2017}

\usepackage[utf8]{inputenc} 
\usepackage[T1]{fontenc}    
\usepackage[colorlinks=true]{hyperref}       
\usepackage{url}            
\usepackage{booktabs}       
\usepackage{amsfonts}       
\usepackage{nicefrac}       
\usepackage{microtype}      
\usepackage{xcolor}
\usepackage{wrapfig}
\usepackage{graphicx}
\usepackage{amssymb}
\usepackage[ruled]{algorithm2e}
\usepackage{authblk}

\definecolor{dgreen}{rgb}{0.0,0.6,0.0}
\hypersetup{citecolor=red} 
\hypersetup{urlcolor=black}
\hypersetup{linkcolor=dgreen} 
  
\def\E{\mathbb{E}}

\newcommand{\Cov}{\mathrm{Cov}}
\newcommand{\Std}{\mathrm{Std}}

\title{Better Text Understanding\\ Through Image-To-Text Transfer}

\author[1]{Karol Kurach}
\author[1]{Sylvain Gelly}
\author[1]{Michal Jastrzebski}
\author[1, 2]{\\Philip Haeusser}
\author[1]{Olivier Teytaud}
\author[1]{Damien Vincent}
\author[1]{Olivier Bousquet}
\affil[1]{Google Brain}
\affil[2]{Technische Universität München}

\begin{document}

\maketitle

\begin{abstract}

Generic text embeddings are successfully used in a variety of tasks. However, they are
often learnt by capturing the co-occurrence structure from pure text corpora,
resulting in limitations of their ability to generalize. In this paper, we explore models that incorporate visual
information into the text representation. Based on comprehensive ablation
studies, we propose a conceptually simple, yet well performing architecture. It
outperforms previous multimodal approaches on a set of well established
benchmarks. We also improve the state-of-the-art results for image-related text
datasets, using orders of magnitude less data.

\end{abstract}

\section{Introduction}

\begin{wrapfigure}[18]{R}{0.5\textwidth}
\centering
\vspace{-\intextsep}
\includegraphics[scale=0.7]{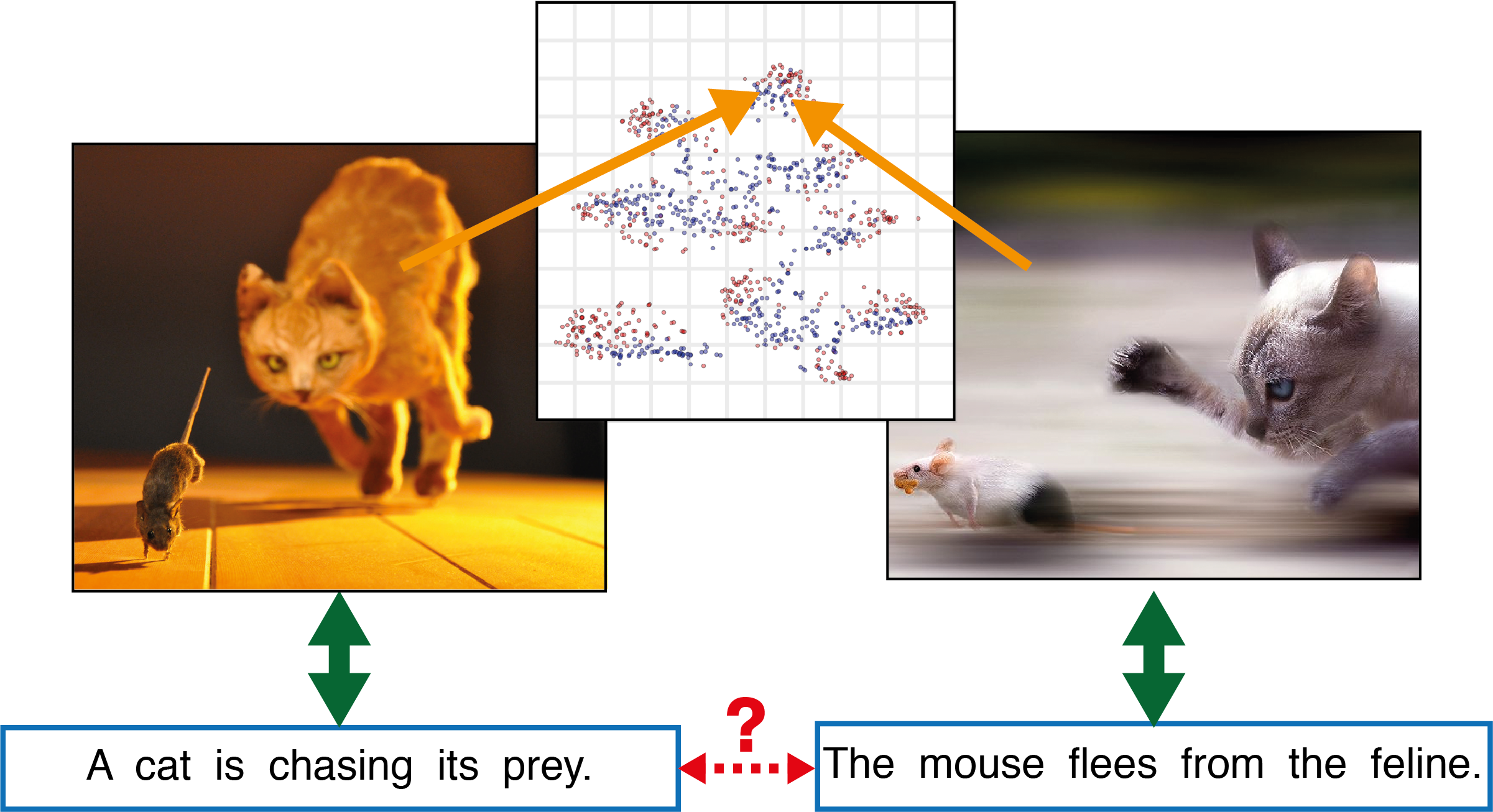}
  \caption{Two images are close in the visual space, which can be quantified via
           a CNN. Their descriptions convey the same concept, yet using an
           entirely different vocabulary.
           Our method improves the understanding of text by
           leveraging knowledge about visual properties of corresponding images.
           Photo credit: \cite{teaser_credit_left, teaser_credit_right}}
\label{fig:circuit_2}
\end{wrapfigure}

The problem of finding good representations of text data is a very actively
studied area of research. Many models are able to learn representations directly
by optimizing the end-to-end task in a supervised manner. This, however, often
requires an enormous amount of labeled data which is not available in many
practical applications and gathering such data can be very costly.  A common solution
that requires an order of magnitude less labeled examples is to reuse
pre-trained embeddings.

A large body of work in this space is focused on training embeddings from pure text
data. However, there are many types of relations and co-occurrences that are
hard to grasp from pure text. Instead, they appear naturally in other modalities, such as images.
In particular, from a similarity measure in the image space and pairs of images and sentences,
a similarity measure for sentences can be induced, as illustrated in \autoref{fig:circuit_2}.

In this work, we study how to build sentence embeddings from
text-image pairs that are good in terms of sentence similary metrics. This extends previous
works, for example \cite{Lazaridou15} or \cite{MaoPinterest16}.

We propose a conceptually simple, but well performing model that we call
Visually Enhanced Text Embeddings (VETE). It takes advantage of visual
information from images in order to improve the quality of sentence embeddings.
This model uses simple ingredients that already exist and combines them
properly.  Using a pre-trained Convolutional Neural Network (CNN) for the image embedding, the sentence
embeddings are obtained as the normalized sum of the word embeddings. Those are
trained end-to-end to be aligned with the corresponding image embeddings and not
aligned with mismatching pairs, optimizing the Pearson correlation.

Despite its simplicity, the model significantly outperforms pure-text models and the best
multimodal model from \cite{MaoPinterest16} on a set of well established text
similarity benchmarks from the SemEval competition \cite{semeval}. In particular, for
image-related datasets, our model matches state-of-the-art results with substantially
less training data. These results indicate that exploring image data can
significantly improve the quality of text embeddings and that incorporating images as a source of information can result in text representations which
effectively captures visual knowledge. We also conduct a detailed ablation study
to quantify the effect of different factors on the embedding quality in the image-to-text
transfer setup. 

In summary, the contributions of this work are:

\begin{itemize}
  \item We propose a simple multimodal model that outperforms previous
    image-text approaches on a wide variety of text similarity tasks.
    Furthermore, the proposed model matches state-of-the-art results on 
    image-related SemEval datasets, despite being trained with substantially
    less data.

  \item We perform a comprehensive study of image-to-text transfer, comparing
    different model types, text encoders, loss types and datasets.

  \item We provide evidence that the approach of learning sentence embeddings directly
    outperforms methods that learn word embeddings and then combine them.

\end{itemize}

\section{Related Work}

Many works study the use of multimodal data, in particular pairs of images and
text. Most of them explore how these pairs of data can be leveraged for tasks
that require knowledge of both modalities, like captioning or image retrieval
\cite{barnard2003matching, jia2011learning, kiros14, Lazaridou15,
wang2016learning, karpathy2015deep, vendrov2015order}.  

While this line of work is very interesting as common embeddings can directly be
applied to captioning or image retrieval tasks, the direct use of text
embeddings for NLP tasks, using images only as auxiliary training data, is less
explored.

\cite{Lazaridou15} propose to extend the skip-gram algorithm \cite{mikolov2013linguistic} to incorporate image data. \cite{hill2014learning} also took a similar approach before. In the original skip-gram algorithm, each word embedding is optimized to increase the likelihood of the neighboring words conditioned on the center word. In addition to predicting contextual words, \cite{Lazaridou15}'s models maximize the similarity between image and word embeddings. More precisely:
\[
 L_{ling}(w_t) = \sum_{-c \leq j \leq c, j\neq0} \log p(w_j | w_t)
\]

where $p(w_j | w_t)$ is given by a softmax formulation:
\[
 p(w_j | w_t) = \frac{e^{f(w_j) \cdot f(w_t)}}{\sum_{w} e^{f(w) \cdot f(w_t)}}
\]

$f(w)$ is the embedding vector of the word $w$ and $v \cdot v'$ is the dot product between the vectors $v$ and $v'$.
To inject visual information, \cite{Lazaridou15} add a max margin loss between the image embedding and the word embedding:

\[
 L_{vision}(w_t) = \E_{w'} \max(0, \gamma - \cos(f(w_t), g(w_t)) + \cos(f(w_t), g(w')))
\]

with $g(w)$ being the average embedding of the images paired with the word $w$.

They show that they can augment word embeddings learnt from large text sources
with visual information and in addition to image labeling and retrieval, they
show that those embeddings perform better on word similarity metrics.

\cite{kiros14} also use a max margin loss to co-embed images and corresponding
text. For the text embeddings, they use different models depending on the task.
For the image captioning or image retrieval task, they employ an LSTM encoder.
To explore the resulting word embedding properties, in particular arithmetics,
they use a simpler word embedding model. Some interesting arithmetical
properties of the embeddings are demonstrated, like ``image of a blue car''
- ``blue'' + ``red'' leading to an image of a red car. However, there is no
quantitative evaluation of the quality of the text embeddings in terms of text
similarity.

Some more recent works investigate phrase embeddings trained with visual signals
and their quality in terms of text similarity metrics. For example,
\cite{MaoPinterest16} use an Recurrent Neural Network (RNN) as a language model in order to learn word
embeddings which are then combined to create a phrase embedding. They propose
three models. For their model A, they use a similar setup as the captioning
model from \cite{Vinyals_2015_CVPR}, with an RNN decoder conditioned on
a pre-trained CNN embedding. The RNN (GRU in that case) reads the text, trying
to predict the next token. The initial state is set to a transformation of the
last internal layer of a pre-trained VGGNet \cite{simonyan2014very}. Let that
vector be $v_{image}$. Model B tries to match the final RNN state with
$v_{image}$. Finally, model C develops the multimodal skip-gram
\cite{Lazaridou15} by adding an additional loss measuring the distance between
the word embeddings and $v_{image}$. The authors' experiments show that model
A performs best and we use that model as a baseline in our experiments.

\section{VETE Models}

\begin{figure}[t]
\centering
\includegraphics[width=\textwidth]{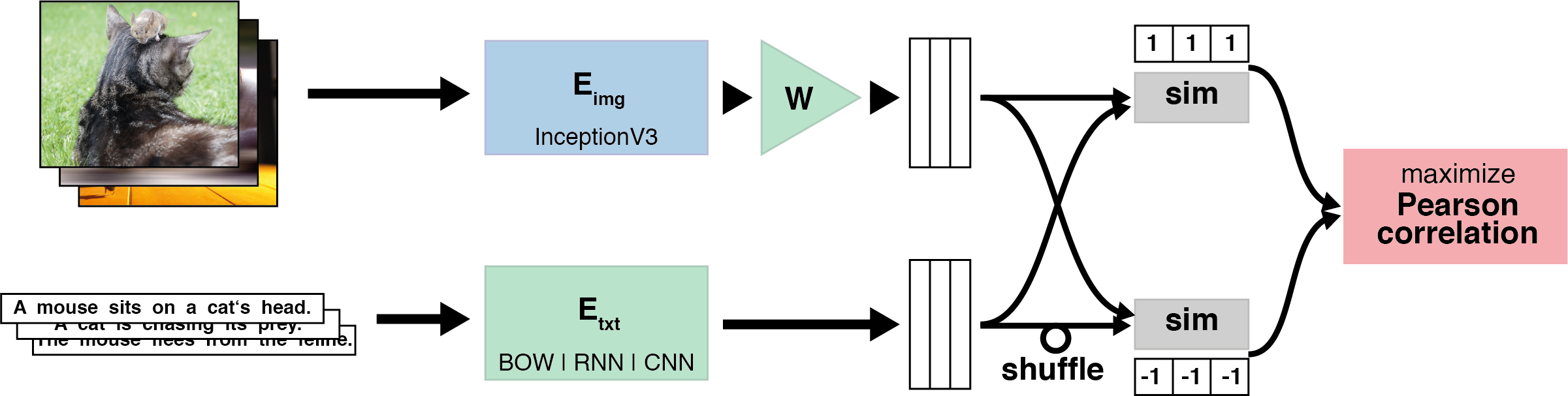}
  \caption{An overview of the VETE model. Images are fed into a pre-trained CNN ($E_{img}$).
  Their representation is then transformed via $W$ to match the dimension of the
  sentence embeddings.
  The sentences are encoded via a text embedding model 
  ($E_{txt}$). Finally, the embeddings are paired in two ways: matching pairs
  and incorrect pairs. For each set of pairs, we compute the similarity.
  The loss signal comes from the Pearson correlation, which should be $1$ for
  matching pairs and $-1$ for incorrect pairs.
  Only green shaded modules are trained.
  Photo credit: \cite{graph_image, teaser_credit_left, teaser_credit_right}}
\label{fig:models_graph}
\end{figure}

Our setup aims at directly transferring knowledge between image and text
representations and the main goal is to find reusable sentence embeddings.
We make direct use of paired data, consisting of pictures and text describing them.
We propose a model consisting of two separate encoders - one for images and another one for text.
An overview of the archiecture is presented in \autoref{fig:models_graph}.

For the text encoder, we consider three families of models which combine words into text
representations. For the bag-of-words (\emph{BOW}) model, the sentence embedding is simply a normalized sum
of vectors corresponding to the individual words. For the \emph{RNN} model, we create a stacked
recurrent neural network encoder (LSTM or GRU based). Finally, for the \emph{CNN} model,
the encoder includes a convolutional layer followed by a fully connected layer,
as described in \cite{kim2014cnn}.

For encoding images, we use a pre-trained \emph{InceptionV3} network \cite{szegedy2016rethinking} which provides a $2048$-dimensional feature vector
for each image in the dataset (images are rescaled and cropped to
$300$ x $300$ px).

Let $E_{img}(I)$ denote the $2048$-dimensional embedding vector for an image $I$ and
$E_{txt}(S)$ the $N$-dimensional embedding for sentence $S$ produced by
$txt \in \{BOW, RNN, CNN\}$.
Throughout this paper, we will refer to the cosine similarity of two vectors $v_1, v_2$ as $sim(v_1, v_2) = \frac{v_1 \cdot v_2}{\|v_1\| \|v_2\|}$.
Informally speaking, our training goal is to maximize $sim(E_{img}(I), E_{txt}(S))$,
when sentence $S$ is paired with (i.e. describes) image $I$, and minimize this
value otherwise.
\newpage
Formally, let $V$ be the vocabulary, $N$ the embedding dimensionality and $txt \in \{BOW, RNN, CNN\}$ the text encoding model.
Let's define an affine transformation $W \in \mathbb{R}^{2048 \times N}$ that transforms the 2048-dimensional image embeddings to $N$ dimensions.
Our learnable parameters consist of a word embedding matrix $\in \mathbb{R}^{|V| \times N}$, the internal parameters of the $txt$ model (when $txt$
is an $RNN$ or $CNN$ encoder), as well as the transformation matrix $W$.
For each batch of image-sentence pairs of the form $(I_1, S_1), ..., (I_B, S_B)$, we randomly shuffle
the sentences $S_1,..., S_B$, and add the following "incorrect" pairs to the batch
$(I_1, S_{\sigma(1)}),..., (I_B, S_{\sigma(B)})$, with $\{\sigma(1),..,\sigma(B)\}$ a random permutation of $\{1,...,B\}$. If we denote
$$ sim(I_i, S_j) := sim(W E_{img}(I_i), E_{txt}(S_j))$$ then our
training goal is to
maximize the Pearson correlation $\rho(x,y) := \frac{\Cov(x,y)}{\Std(x)\Std(y)}$
between the vectors
$$[\underbrace{sim(I_1, S_1), ..., sim(I_B, S_B)}_{B \mbox{ \small correct pairs}},\underbrace{sim(I_1, S_{\sigma(1)}), ..., sim(I_B, S_{\sigma(B)})}_{B \mbox{ \small wrong pairs}}]$$
and
$$ [\underbrace{1, 1, ..., 1}_{B\mbox{ \small positive labels}}, \underbrace{-1, -1, ..., -1}_{B\mbox{ \small negative labels}}]. $$

We will denote models trained this way VETE-BOW, VETE-RNN and VETE-CNN, respectively.

\section{Experiments}

\subsection{Training Datasets}

In our experiments we consider three training datasets: MS COCO, SBU and Pinterest5M.
They are described in detail below. One important note is that we modify datasets that
contain multiple captions per image (MS COCO and Pinterest5M) to keep only one
caption.  This was done to prevent the network from ``cheating'' by using the
image feature vector only as a way of joining text pairs.  To the best of our
knowledge, we are the first to notice this issue in the evaluation of the
quality of multimodal image-text models. It is known that training similarity
models directly on text-text pairs yields good results \cite{wieting2015towards} but
here we want to investigate only the effect of knowledge transfer from images.

\begin{figure}[t]
\centering
\includegraphics[width=\textwidth]{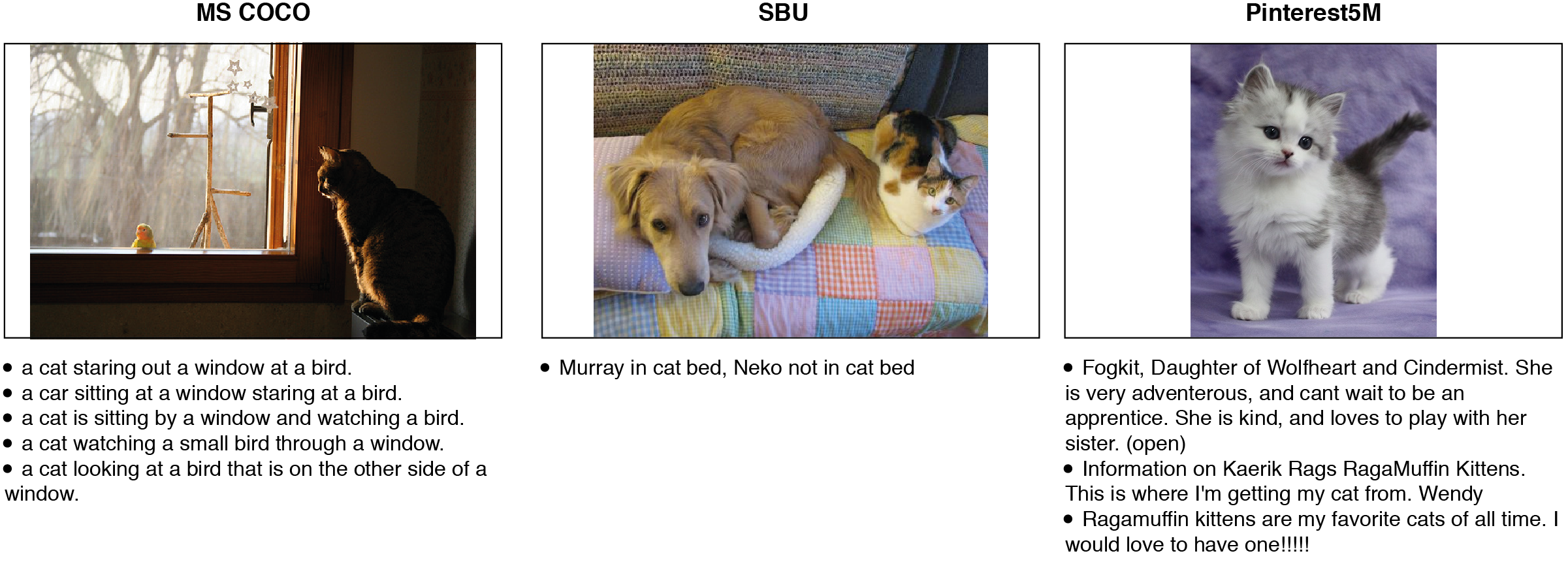}
  \caption{Examples from the three datasets. Shown are images and all provided
  captions. Notice how the language varies from formal descriptions geared
  towards a general audience (MS COCO) to more informal posts (SBU,
  Pinterest5M).}
\label{fig:datasets}
\end{figure}

\paragraph{MS COCO}
The MS COCO dataset \cite{coco2014} contains $80$ image categories. For each image five
high-quality captions are provided. We use the MS COCO 2014 dataset using the same
train/validation/test split as the im2txt \cite{im2txt_impl} Tensorflow implementation of \cite{im2txt_paper}.
Initially, our
train/validation/test sets contain 586k/10k/20k examples, respectively. Then, we filter the sets to keep only
one caption per image, so the "text part" of our final datasets is five times smaller.

\paragraph{SBU}
The Stony Brook University dataset\cite{Ordonez:2011:im2text} consists of 1M image-caption pairs collected
from Flickr, with only one caption per image. We randomly split this dataset into train/validation/test
sets with sizes 900k/50k/50k, respectively.

\paragraph{Pinterest5M}

The original Pinterest40M dataset \cite{MaoPinterest16} contains $40$M images. 
However, only $5$M image urls were released at the time of this submission.
Unfortunately, some images are no longer available so we were able to collect
approx. $3.9$M images from this dataset. For every image we keep only
one caption. Then, we randomly split the data into $3.8$M/$50$k/$50k$ train/validation/test sets, respectively.

The training data in all datasets is lowercased and tokenized using the Stanford
Tokenizer \cite{stanfordtokenizer}. We also wrap all sentences with "<S>" and
"</S>" marking the beginning and the end of the sentence.

\subsection{Hyperparameter selection and training}\label{sec:hyperparam_selection}

The performance of every machine learning model highly depends on the choice of its hyperparameters.
In order to fairly compare our approach to previous works, we follow the
same hyperparameter search protocol for all models. We choose the average score on the SemEval 2016 (c.f. \autoref{sec:evaluation}) as the validation metric and refer to this as ``avg2016''.

\begin{algorithm}[H]
\For{i=1,2,\dots,100}{
  Sample a set of hyperparameters in the allowed ranges\;
  Run training and evaluate on ``avg2016''\;
  }
  Report the results on all benchmarks of the model that has the highest score on ``avg2016''\;

 \caption{Protocol for hyperparameter search.}
\end{algorithm}

If a hyperparameter has a similar meaning in two 
models (e.g., learning rate, initialization scale, lr decay, etc.), the ranges searched
were set to be the same. Additionally, we ensured that the parameters
reported by the authors are included in the ranges. 

In all models, we train using the Adam optimizer\cite{kingma2014adam}, for $10$ (MS COCO, SBU)
or $5$ epochs (Pinterest5M). The final embeddings have size of $128$.
For all VETE models we use the Pearson loss (see ablation study in \autoref{sec:ablation_loss}).

\begin{table}[t]
\centering
\begin{tabular}{|c|c|c|c|c|c|c|}
  \hline
  \textbf{Model} & \textbf{imagess2014} & \textbf{images2015} & \textbf{COCO-Test} & \textbf{Pin-Test} & \textbf{avg2014} & \textbf{avg2015} \\
  \hline
  Word2Vec & $0.466$ & $0.441$ & $0.379$ & $0.383$ & $0.343$ & $0.367$ \\
  \hline
  PureTextRnn & $0.662$ & $0.692$ & $0.705$ & $0.484$ & $0.517$ & $0.568$ \\
  \hline
  PinModelA & $0.671$ & $0.683$ & $0.709$ & $0.536$ & $0.493$ & $0.573$ \\
  \hline
  \hline
  VETE-RNN & $0.838$ & $0.835$ & $0.901$ & $0.549$ & $0.538$ & $0.587$ \\
  \hline
  VETE-CNN & $0.808$ & $0.773$ & $\textbf{0.911}$ & $0.528$ & $0.435$ & $0.486$ \\
  \hline
  VETE-BOW & $\textbf{0.861}$ & $\textbf{0.855}$ & $0.894$ & $\textbf{0.579}$ & $\textbf{0.579}$ & $\textbf{0.622}$ \\
  \hline
\end{tabular}
\newline
\caption{Results of models trained only on MS COCO data with one sentence per image.}\label{fig:image_on_coco}
\end{table}

\subsection{Evaluation}\label{sec:evaluation}

Our goal is to create good text embeddings that encode knowledge from corresponding
images.  To evaluate the effect of this knowledge transfer from images to text, we use a set
of textual semantic similarity datasets from the SemEval $2014$ and $2015$
competitions\cite{semeval}.
Unfortunately, we could not compare our models directly on the Gold RP10K dataset
introduced by \cite{MaoPinterest16} as it was not publicly released.

We also use two additional custom test sets: \emph{COCO-Test} and
\emph{Pin-Test}. These were created from the MS COCO and Pinterest5M test
datasets, respectively, by randomly sampling $1,000$ semantically related captions
(from the same image) and $1,000$ non-related captions from different images.
As opposed to the SemEval datasets, the similarity score is binary in this case.
The goal was to check the performance on the same task as SemEval but
with data from the same distribution of words as our training data.

For every model type we select the best model according to the average score
on the SemEval $2016$ datasets.  Then, we report the results on all other test
datasets.

\subsection{Results}

\autoref{fig:image_on_coco} presents the scores obtained by models trained
only on MS COCO datasets. This allows us to fairly compare only the algorithms,
not the data used. In \autoref{sec:ablation_dataset}, we analyze the robustness of our methods on two additional datasets (SBU and Pinterest5M).

As a direct comparison, we implement $Model A$ as described in
\cite{MaoPinterest16}, which we refer to as \emph{PinModelA}. Our implementation
uses a pre-trained InceptionV3 network for visual feature extraction, as in the
VETE models. To understand the impact of adding information from images to text
data, we also evaluate two models trained purely on text:
\begin{itemize}
\item \textbf{RNN-based language model} This model learns sentence embeddings via
an RNN based language model. It corresponds to the PureTextRnn baseline from \cite{MaoPinterest16}.
\item \textbf{Word2Vec} We trained Word2Vec word embeddings \cite{rehurek_lrec} where the corpus
consists of sentences from MS COCO.
\end{itemize}

All VETE models outperform pure-text baselines and PinModelA. Similarly
to \cite{wieting2015towards}, we observed that RNN-based encoders are
outperformed by a simpler BOW model. We also show that this holds for CNN-based
encoders. It is worth noting that it is mostly a domain adaptation issue, as both
RNN and CNN encoders perform better than BOW on \emph{COCO-Test}, where the data has
the same distribution as the training data. We analyze the effect of changing
the text encoder in \autoref{sec:ablation_encoder}.

To put our results in context, \autoref{fig:image_open} compares them to other methods trained with much larger corpora.
We used word embeddings obtained from three methods:
\begin{itemize}
   \item \textbf{Glove}: embeddings proposed in \cite{pennington2014glove}, trained on a Common Crawl dataset with 840 billion tokens.
   \item \textbf{M-Skip-Gram}: embeddings proposed in \cite{Lazaridou15}, trained on Wikipedia and a set of images from ImageNet.
   \item \textbf{PP-XXL}: the best embeddings from the  \cite{wieting2015towards}, trained on 9M phrase pairs from PPDB.
\end{itemize}

For all three approaches, we consider two versions that differ in the vocabulary
allowed at inference time.
One experiment was done with a vocabulary
restricted to MS COCO (marked with ``R'') and
the non-restricted version (``NR'') where we use the whole vocabulary for
given embeddings. The vocabulary size has a significant impact
on the final score for the Pinterest-Test benchmark, where $16.5$\% of all
tokens are not in the MS COCO vocabulary. That means that $97.6$\% of all sentences
have at least one missing token.

Finally, we also include the best results from the SemEval competition, where
available. Note that those were obtained from heavily tuned and more complex
models, trained without any data restrictions. Still, our VETE model is able to match
their results.

\begin{table}
\centering
\begin{tabular}{|c|c|c|c|c|}
  \hline
  \textbf{Model} & \textbf{images2014} & \textbf{images2015} & \textbf{COCO-Test} & \textbf{Pin-Test} \\
  \hline
  Glove (R) & $0.624$ & $0.686$ & $0.668$ & $0.422$ \\
  \hline
  Glove (NR) & $0.625$ & $0.688$ & $0.667$ & $0.471$ \\
  \hline
  PinModelA & $0.671$ & $0.683$ & $0.708$ & $0.536$  \\
  \hline
  M-Skip-Gram (R) & $0.764$ & $0.767$ & $0.784$ & $0.608$ \\
  \hline
  M-Skip-Gram (NR) & $0.765$ & $0.767$ & $0.784$ & $\textbf{0.654}$ \\
  \hline
  PP-XXL (R) & $0.802$ & $0.831$ & $0.770$ & $0.609$ \\
  \hline
  PP-XXL (NR) & $0.804$ & $0.833$ & $0.770$ & $0.638$ \\
  \hline
  Best SemEval & $0.834$ & $\textbf{0.871}$ & N/A & N/A  \\
  \hline
  \hline
  VETE-BOW (our) & $\textbf{0.861}$ & $0.855$ & $\textbf{0.894}$ & $0.579$   \\
  \hline
\end{tabular}
\newline
\caption{Comparison of models on image-related text datasets. VETE and PinModelA
were trained only on MS COCO.}\label{fig:image_open}
\end{table}

\section{Ablation studies}

To analyze the impact of different components of our architecture, we perform
ablation studies on the employed text encoder, loss type and training dataset. We also
investigate the effect of training on word or sentence level.  In all cases, we
follow a similar protocol as in \autoref{sec:hyperparam_selection}:

\begin{algorithm}[H]
Randomly generate $100$ sets of combinations for all hyperparameters.\;
\For{\upshape Hyperparameter \textbf{p} (e.g, ``loss type'')}{
  \For{\upshape \textbf{v} in the allowed range of values for \textbf{p}}{
    Run training using the 100 sets of hyperparameters, keeping \textbf{p}=\textbf{v} fixed.\;
  }
  Choose the best one based on ``avg2016'' validation metric, and report the scores.\;
}
 \caption{Protocol for hyperparameter ablation study.}
\end{algorithm}

\subsection{Encoder}\label{sec:ablation_encoder}

We study the impact of the different text encoders on the VETE model. The results are
summarized in \autoref{fig:table_encoder}.
``RNN-GRU'' and ``RNN-LSTM'' denote RNN encoders with GRU~\cite{gru} and
LSTM~\cite{lstm} cells, respectively. For BOW, we try two options: either we use
the sum or the mean of word embeddings.
Both bag-of-words encoders perform better than RNN encoders, although RNNs are
slightly better on the test data which has the same distribution as the training data.

\begin{table}[!htb]
  \begin{minipage}[t]{0.67\linewidth}
    \vspace{-0.81in}
    \scalebox{0.89}{
      \begin{tabular}{|c|c|c|c|c|}
        \hline
        \textbf{Encoder} & \textbf{images2014} & \textbf{images2015} & \textbf{COCO-Test} & \textbf{Pin-Test} \\
        \hline
        RNN-GRU & $0.834$ & $0.821$ & ${\bf 0.906}$ & $0.507$ \\
        \hline
        RNN-LSTM & $0.838$ & $0.835$ & $0.901$ & $0.549$ \\
        \hline
        BOW-SUM & $0.860$ & $0.853$ & $0.898$ & $0.573$ \\
        \hline
        BOW-MEAN & ${\bf 0.861}$ & ${\bf 0.855}$ & $0.894$ & ${\bf 0.579}$ \\
        \hline
      \end{tabular}
    }
    \hspace{0.05in}
    \vspace{0.28in}
    \caption{Results of applying different text encoders to the VETE model. The
      training data is MS COCO, and the RNN-based models learned to model this distribution
      better. However, BOW generalizes better to other datasets.}\label{fig:table_encoder}
  \end{minipage}%
  \begin{minipage}[t]{0.02\linewidth}
   \textcolor{white}{.}
  \end{minipage}%
  \begin{minipage}{0.3\linewidth}
    \scalebox{0.89}{
      \begin{tabular}{|c|c|}
        \hline
        \textbf{Loss type} & \textbf{Avg score} \\
        \hline
        Covariance & $0.594$ \\
        \hline
        SKT$_{.2}$ & $0.616$ \\     
        \hline
        SKT$_1$ & $0.730$ \\        
        \hline
        Rank loss & $0.788$ \\
        \hline
        SKT$_5$ & $0.791$ \\        
        \hline
        Pearson & ${\bf 0.797}$ \\
        \hline
      \end{tabular}
    }
    \hspace{0.05in}
    \vspace{0.06in}  
    \caption{Comparison of different loss types with the VETE-BOW model.}\label{fig:ablation_loss}
  \end{minipage}
\end{table}

\subsection{Loss type}\label{sec:ablation_loss}
In this section, we describe various loss types that we trained our model with.
Consider two paired variables $x$ (similarity score between two embeddings) and
$y \in \{-1, 1\}$. Then, the sample sets $(x_1,\dots,x_n)$ and $(y_1,\dots,y_n)$ stand for $n$ corresponding realizations of $x$ and $y$.

\begin {itemize}
\item \textbf{Covariance}: $\Cov(x,y)$.
\item \textbf{The Pearson correlation $\rho$}: measures the linearity of the link between two variables, estimated on a sample; it is defined as $\rho(x,y) = \frac{\Cov(x,y)}{\Std(x)\Std(y)}$.
\item \textbf{Surrogate Kendall $\tau$}: The Pearson correlation takes into account only
linear dependencies. To mitigate this, we experimented with the Kendall correlation $\tau$
which is only rank-dependent. Unfortunately, it is not differentiable. We therefore used its
differentiable approximation: \textbf{$SKT_\alpha$} \cite{skt} defined as
$SKT_\alpha(x,y)=\frac{ \sum_{i,j} \tanh \left( \alpha (x_i-x_j)(y_i-y_j) \right) }{n(n-1)/2}$
for some $\alpha>0$.
\item \textbf{Rank loss}: Another cost function is the pairwise ranking loss.
We follow closely the definition in \cite{kiros14}.
\end{itemize}
\autoref{fig:ablation_loss} compares the effects of the various losses.

\subsection{Dataset}\label{sec:ablation_dataset}

We study the effect of the training dataset. The results of training the model
on MS COCO, SBU and Pinterest5M dataset are presented in 
\autoref{fig:datasets_table}.  Each cell of the table contains the average score
of $4$ evaluation datasets (images2014, images2014, COCO-Test, Pin-Test). 
The quality of image captions varies significantly between the datasets,
as can be seen in \autoref{fig:datasets}. However, we conclude that the
relation between the models is preserved, that is: regardless of the dataset used for
training, PinModelA is always worse than VETE-RNN, which in turn is worse than
VETE-BOW.

\begin{table}[h]
\centering
\begin{tabular}{|c|c|c|c|c|}
  \hline
  \textbf{Train Dataset} & \textbf{Word2Vec} & \textbf{PinModelA} & \textbf{VETE-RNN} & \textbf{VETE-BOW} \\
  \hline
  MS COCO & $0.417$ & $0.650$ & $0.780$ & ${\bf 0.797}$ \\
  \hline
  SBU & $0.413$ & $0.632$ & $0.737$ & ${\bf 0.775}$ \\
  \hline
  Pinterest5M & 0.408 & $0.609$ & $0.753$ & ${\bf 0.803}$ \\
  \hline
\end{tabular}
\newline
\caption{Comparison of average test scores when training on different datasets.}\label{fig:datasets_table}
\end{table}

\subsection{Sentence-level vs word-level embedding}

Previous methods for transferring knowledge from images to text focused on
improving the word-level embeddings. A sentence representation could then be
created by combining them.  In our work, we learn sentence embeddings as
a whole, but the best performing text encoder turned out to be BOW. This raises
the following question: could the model perform equally well if we train it on
word-level, and then only combine word embeddings during inference?  The
comparison of these two approaches is presented in
\autoref{fig:ablation_sentence_word} which clearly shows the benefit of
sentence-level training. This effect should be studied further, but while separately training
word embeddings forces each of them to be close to the corresponding images,
training at the sentence level gives the opportunity to have the word embeddings become complementary,
each of them explaining a part of the image, and capturing co-occurences.

\begin{table}[h!]
\centering
\begin{tabular}{|c|c|c|c|c|}
  \hline
  \textbf{Model} & \textbf{images2014} & \textbf{images2015} & \textbf{COCO-Test} & \textbf{Pin-Test} \\
  \hline
  Word-level & $0.576$ & $0.617$ & $0.675$ & $0.371$ \\
  \hline
  Sentence-level & $\textbf{0.861}$ & $\textbf{0.855}$ & $\textbf{0.894}$ & $\textbf{0.579}$ \\
  \hline
\end{tabular}
\newline
\caption{Comparison of training on a word-level vs sentence-level.}\label{fig:ablation_sentence_word}
\end{table}

\section{Conclusion}
We studied how to improve text embeddings by leveraging multimodal datasets, using a pre-trained image model and paired text-image datasets. We showed that VETE, a simple approach which directly optimizes phrase embeddings to match corresponding image representations, outperforms previous multimodal approaches which are sometimes more complex and optimize word embeddings as opposed to sentence embeddings. We also showed that even for relatively complex similarity tasks at sentence levels, our proposed models can create very competitive embeddings, even compared to more sophisticated models trained on orders of magnitude more text data, especially when the vocabulary is related to visual concepts. 

To our initial surprise, state-of-the-art encoder models, like LSTMs, performed significantly worse than much simpler encoders like bag-of-word models. While they achieve better results when evaluated on the same data distribution, their embeddings do not transfer well to other text distributions. General embeddings need to be robust to distribution shifts and applying such techniques can probably further improve the results.

Using a multimodal approach in order to improve general text embeddings is under-explored and we hope that our results motivate further developments. For example, the fact that the best models are very simple suggests that there is a large headroom in that direction.

\newpage

\bibliographystyle{abbrv}

\bibliography{multimodal}
\end{document}